\documentclass{article}
\usepackage{spconf,amsmath,graphicx,hyperref}
\usepackage{caption}
\usepackage{mathtools}
\usepackage{amsfonts} 
\usepackage{booktabs}
\usepackage{subcaption}

\title{T-GEMs: Text-Guided Exit Modules for decreasing clip image encoder}
\name{Alberto Presta$^{1*}$\thanks{*Corresponding author. Email: a.presta@samsung.com}, Grzegorz Stefanski$^{1}$, Michal Byra$^{1,2}$, Krzysztof Arendt$^{1}$}
\address{$^{1}$Samsung AI Center Warsaw  \quad $^{2}$Institute of Fundamental Technological Research, PAS, Warsaw}
\begin{document}
%
\maketitle
\begin{abstract}
Multimodal deep neural networks enhance deep comprehension by integrating diverse data modalities. Data from different modalities are typically projected into a shared latent space for similarity computation, but this process is resource intensive due to large image encoders and equal processing of test data during prediction. Early exit methods reduce computational load by utilizing intermediate layers, saving time and memory. However, developing such methods is challenging for multimodal data like image-text pairs. This study investigates the semantic content distributions present in intermediate layers of encoders such as CLIP, which can be derived from textual descriptions. We introduce Text-Guided Exit Modules (T-GEMs) and a rate-based regularizer to control encoder usage costs while maintaining cross-modal understanding performance.
\end{abstract}

\section{introduction and related works}

\vspace{-0.25cm}
Large-scale multimodal (MM) models such as CLIP \cite{Radford2021LearningTV}, ALIGN \cite{Jia2021ScalingUV}, and Florence \cite{Yuan2021FlorenceAN} process image and text inputs independently and align them in a shared embedding space using contrastive learning. While CLIP is efficient at scale, its inference process is static: each input is passed through the full transformer stack, regardless of complexity. 

To enhance computational efficiency, the early exit (EE) strategy has been introduced as a viable approach, either dynamically, where the computational cost is tailored to individual test samples, or statically, wherein the reduction in computation is uniformly applied across all samples.
First introduced in BranchyNet \cite{Teerapittayanon2016}, EE methods add auxiliary classifiers at intermediate layers, allowing models to stop computation when sufficiently confident. This idea has been extended with other architectures \cite{Kaya2018ShallowDeepNU,huang2018multiscale}.
Static methods to perform EE with deterministic and fixed rules have been introduced \cite{liu2021faster, sun2022simple}.
Alternatively, dynamic confidence-based approaches have been designed for residual and transformer models \cite{Yin2021AViTAT, xin2020deebert, Elbayad2020Depth, liao2021global,li2020cascadebert, eyzaguirre2021dact}, as long as adaptive inference time optimal policy \cite{ilhan2024adaptive, schuster2021consistent}.
These methods aim to reduce computation while maintaining accuracy.

EE methods in MM models are still largely underexplored; recent work explored adaptive inference , using techniques such as token pruning and conditional computation to reduce cost. 
Token Merging \cite{bolya2023token} eliminates redundant visual tokens, and multiple exiting (MuE) \cite{Tang2023You, fei2022deecap} enables dynamic layer skipping based on cross-modal similarity.
However, these approaches focus on fusion-based models that process modalities jointly, rather than on contrastive dual-encoder models like CLIP, where the separate modality encoders and late fusion make adaptive inference more challenging; In contrast to our approach, such methods do not incorporate information from the text encoder for EE tasks on the visual encoder, keeping them separate.

Even if we do not develop a dynamic EE method where each test sample has been processed individually, to our knowledge, this is the first study to explore how to exploit intermediate layers of both text and visual encoder in contrastive dual-encoder models such as CLIP in order to perform static EE. 
Our contributions are as follows: (i) analysis of how intermediate representations vary across classes with respect to the text embedding; (ii) introduction of the \emph{class-rate}, a statistic derived from the activation map that correlates with cosine similarity and can be used for training as a regularizer; (iii) design of a module that predicts intermediate image encoder activations based solely on text embeddings, revealing a bounded relationship between modalities; and (iv) training of static EE modules that leverages intermediate activation maps for tasks such as image classification.

\begin{figure*}[t]
  \centering
  \begin{subfigure}[b]{0.30\textwidth}
    \includegraphics[width=\textwidth]{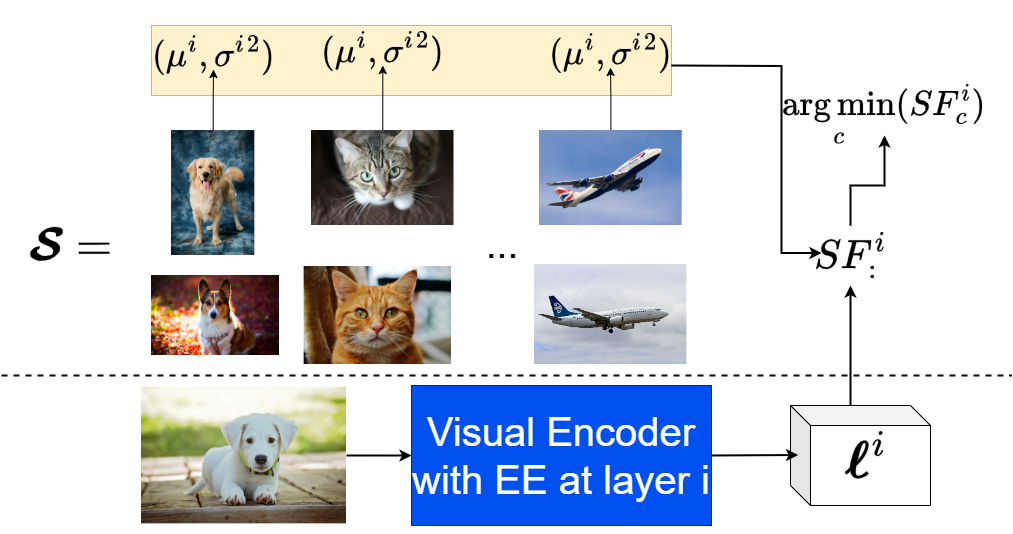}
    \caption{}
    \label{fig1}
  \end{subfigure}
  \hfill
  \begin{subfigure}[b]{0.28\textwidth}
    \includegraphics[width=\textwidth]{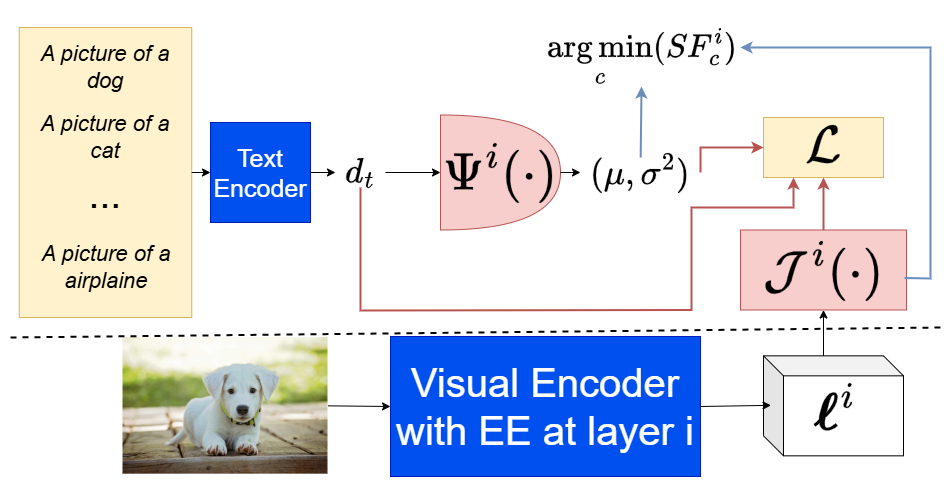}
    \caption{}
    \label{fig2}
  \end{subfigure}
  \hfill
  \begin{subfigure}[b]{0.24\textwidth}
    \includegraphics[width=\textwidth]{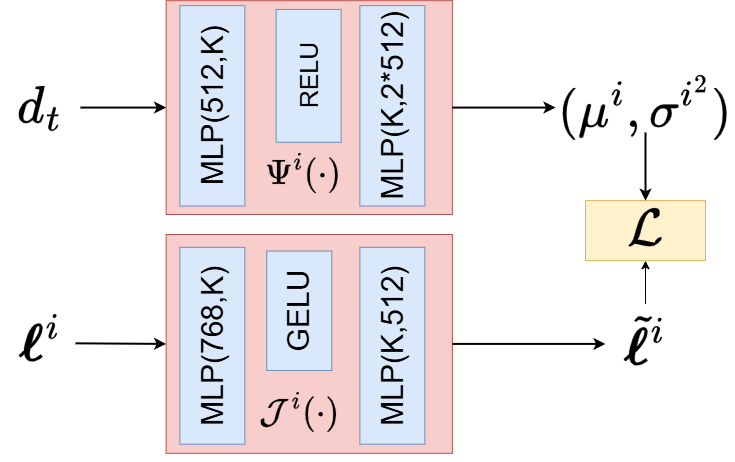}
    \caption{}
    \label{fig3a}
  \end{subfigure}
    \vspace{-0.5cm}
      \caption{(a) Sampling based approach vs (b) Learning based approach to estimate distribution parameters for $\boldsymbol\ell^{i}$. In (b), red and blue arrows represent training and inference phases, respectively. (c) architecture of $\Psi^i$ and $\mathcal{J}^{i}(\cdot)$. $SF$ stands for similarity function, that can be either the cosine similarity or the class-rate. }
  \label{fig:three_images} 
\end{figure*}

\section{ proposed method}
\vspace{-0.3cm}
On top of a MM dual-encoder model like CLIP \cite{Radford2021LearningTV}, we introduced a method that leverages either a small calibration set or text embeddings to utilize intermediate layers for classification, thereby reducing the size of the image encoder.
Given an image $x$ and a text description $d$ corresponding to a specific class $c$, let $\ell^{ij}$ denote the $j$-th neuron output of the $i$-th intermediate layer from which an EE may occur.
Our goal is to analyze individual neuron activation patterns for each class and identify unique posterior probability distributions. Each neuron processes a fixed number of tokens T, consisting of T-1 patch tokens and one class token. To generalize across tasks, we calculate the average of these tokens to derive a single representative value.
Primarily, we aim to estimate the following posterior probability $ p_{x,d \sim c} (\ell^{ij}) $.
We impose a  gaussian prior distribution  since activations from such models are well represented by it \cite{peer2022improving}. 
Therefore, we can simplify our problem to computing the mean and variance of the Gaussian  for each class $c$, obtaining the following distribution
\begin{equation}\label{eq2}
    p_{x,d \sim c} (\ell^{ij} | x,d)  \sim \mathcal{N}_{c}^{ij}(\mu_c^{ij}, \sigma_c^{ij^2}),
\end{equation}
\noindent where $\mu_c^{ij}$ and $\sigma_c^{ij^2}$ stand for the Gaussian distribution mean and variance for class $c$, respectively.  

Once we estimate the Gaussian distribution parameters for each class $c$ and each layer, we can leverage such values to perform classification on intermediate layers, performing EE.
From a test image $x$, we extract its activations from the $i$-th  intermediate layer $\boldsymbol\ell^{i:} \in \mathbb{R}^{N} $ where $N$ is the number of neurons, and for each class $c$ we compute the \emph{class-rate} $ \mathcal{R}_{c}^{i}$ as follows: 

\begin{equation} \label{eq3}
     \mathcal{R}_{c}^{i} = \frac{1}{N} \sum_{j = 1}^{N} \mathcal{R}_{c}^{ij} 
      =  \frac{1}{N} \sum_{j = 1}^{N} \log_{2}(\sigma_c^{ij^2}) + \frac{(\ell^{ij} - \mu_c^{ij})^2}{2\sigma_c^{ij^2}},
\end{equation}    

\noindent where eq. \ref{eq3} is for the Gaussian assumption where we removed constant values.
After computing such value for each class, we can compute the final response 
\begin{equation}\label{eq4}
    y^{i} = \mathop{\arg\min}_{c}(\mathcal{R}_{c}^{i})
\end{equation}

where $ y^{i}$ represents the prediction derived from  layer $i$.
We emphasize that this approach is not tied to a single supervised task, but it can be generally applied to compare inputs with different semantic characteristics.
\\
Roughly speaking, the \emph{class-rate} described by eq. \ref{eq3} is the amount of \emph{surprise} we have by observing $l^{ij}$ and considering class $c$ as ground truth class.
Because of this, taking the minimum in eq. \ref{eq4} means that we are taking the class which leads to the greatest compression of the signal $l^{ij}$; i.e. the one with the least amount of \emph{surprise} when observing $l^{ij}$. 
This concept is fundamental in learning-based signal compression \cite{jamil2023learning}; in this context, it has been utilized to identify semantically guided peculiarities within intermediate layers, enabling the determination of optimal EE points to retain such information.

However, of fundamental importance is how to compute $(\mu_c^{ij}, \sigma_c^{ij^2})$, which is described in the following sections.

\subsection{Sampling based distribution parameters estimation} \label{sb}
The first naive approach to compute $(\mu_c^{ij}, \sigma_c^{ij^2})$, shown in fig. \ref{fig1}, is toward simple sampling, i.e. to rely on a calibration set $\boldsymbol{\mathcal{S}}$ where each semantic class is well represented by a certain number of instances.
Starting from  $\boldsymbol{\mathcal{S}}$, for each class $c$ and layer $i$, we simply consider the intermediate layers $\boldsymbol\ell^{i:}_{c} \in \mathbb{R}^{N} $  and we compute the mean $\mu_c^{ij}$ and the variance $\sigma_c^{ij^2}$ for each neuron $j$.
This approach is the easiest way to extract the Gaussian posterior distribution $ p_{x,d \sim c} $, and it avoids introducing additional parameters and training phase into the system.
However, this approach has two main limitations: i) it necessitates the use of   $\boldsymbol{\mathcal{S}}$ and relies heavily on it, and ii) it fails to establish a connection between textual and image activation maps, failing thus to be multi-modal.

\begin{table*}[ht]
\centering

\caption{Classification accuracy obtained by applying EE after each of the  RB composing CLIP VIT-32-B/VIT-L-14.
SF stands for \emph{similarity function}, i.e. the function we use to perform classification at inference time, while AP stands for additional parameters, i.e. the parameters we add for EE. }\label{tab1}
\resizebox{\textwidth}{!}{%
\begin{tabular}{l ccc cccccccccccc}
\toprule
\textbf{Method} &   \textbf{Loss} &   \textbf{SF} & \textbf{AP} &
\multicolumn{12}{c}{\textbf{CIFAR10 (VIT-B-32 / VIT-L-14)}}  \\
\cmidrule(lr){5-16} 
& & & & RB1 & RB2 & RB3 & RB4 & RB5 & RB6 & RB7 & RB8 & RB9 & RB10& RB11 & RB12 \\
\midrule
Sampling-based & -  & Class-rate  & 0 &   27/20 & \textbf{32}/\textbf{25}  & \textbf{37}/26 & \textbf{41}/\textbf{30} & \textbf{48}/\textbf{35} &\textbf{57}/\textbf{36} & \textbf{61}/\textbf{38} & \textbf{65}/\textbf{41} & \textbf{77}/\textbf{43} & \textbf{85}/\textbf{48} & \textbf{83}/\textbf{54} & \textbf{82}/56 \\
Sampling-based & -  & Cosine & 0 &  \textbf{29}/\textbf{24} & 30/\textbf{25}  & 33/\textbf{29} & 37/\textbf{30} & 42/31&50/33 & 57/36 &62/38 & 70/\textbf{43} & 76/\textbf{48} & 71/\textbf{54}& 76/\textbf{58}\\
\bottomrule
Jumper  & $\mathcal{L}_{s}$  & Cosine  & 164.5/230k & 40/46 & 46/51 & 50/ \textbf{54} & 56/ \textbf{55} & 62/54 & 76/60 & 80/65 & \textbf{84}/69& 86/\textbf{71} & 91/ \textbf{78} & 92/78& 91/83 \\
T-GEMs  + Jumper  & $\mathcal{L}_{r}$   & class-rate & 362/592k & 14/36 & 18/34 & 13/38 & 16/38 & 18/41 & 26/43 & 35/41 & 43/49& 48/45 & 54/48 & 53/65 & 54/51 \\
T-GEMs + Jumper  & $\mathcal{L}_{r} +  \mathcal{L}_{s}$  & class-rate & 362/592k & 38/40 & 44/42 & 48/47 & 48/49 &60/46 & 62/53 & 79/56 & 78/62 & 84/62 &89/67 &83/67&89/77  \\
T-GEMs + Jumper  & $\mathcal{L}_{r} +  \mathcal{L}_{s}$  & Cosine & 362/592k &\textbf{41}/\textbf{48} & \textbf{47}/\textbf{52} & \textbf{55}/53 & \textbf{60}/54 &\textbf{68}/\textbf{55} & \textbf{77}/\textbf{64} & \textbf{81}/\textbf{66} &83/\textbf{70}&\textbf{87}/\textbf{71}& \textbf{92}/77 & \textbf{92}/\textbf{79} & \textbf{91}/\textbf{84}  \\
\bottomrule
CLIP (zero-shot)  & - & -  & - &\multicolumn{12}{c}{\textbf{91.3}/\textbf{96.2}}  \\
\end{tabular}%
}
\end{table*}

\subsection{Learning based  distribution parameters estimation}
The second approach, shown in fig. \ref{fig2}, involves using the text encoder to estimate the distribution parameters without the use of sample images.
Given  the intermediate layer $\boldsymbol{\ell^{i:}}$, this approach consists of a learnable network $\Psi_i$ called T-GEM (\emph{Text-Guided-Exit-Module}) that learns the distribution parameters starting from the output of the text encoder; this approach is described as follows
\begin{equation}\label{eq5}
    \Psi_i(d_t, \phi) = (\mu^{i:}, \sigma^{i:^2}) \in \mathbb{R}^{2\times N} 
\end{equation}
where $d_t$ is the text embedding and $\phi$ represents the learnable parameters.
We assume that the text encoder includes all the necessary information to extract distribution parameters, because of the original contrastive loss used to train the CLIP.
Eq. \ref{eq5} can be compared to hyperprior-based modules used in learned image compression \cite{balle2018variational}, where each channel has been independently compressed.
$\Psi_i$ is optimized by minimizing the rate as follows:

\begin{equation}  \label{loss}
     \mathcal{L}_{r} = -\mathbb{E} [ \log_2(\mathcal{N}(\boldsymbol{\ell^{i:}}|\mu^{i:}, \sigma^{i:^2}) ].
\end{equation}

\noindent The class is not explicitly introduced in eq. \ref{loss}, and the convergence comes naturally with the image-text pairing. 

Our method has been further improved by the introduction of a third module $\mathcal{J}^i$, called \emph{Jumper}, whose major aim is to project the $i$-th intermediate layer $\boldsymbol{\ell^{i:}}$ in a more suitable representation, facilitating the optimization of eq. \ref{loss}.
The goal is to find a space where the rate estimation can have a major discriminatory factor between different classes, in a similar way as it is performed in learned compression.
Our module, i.e. Jumper + T-GEM (Fig. \ref{fig2}) is described as follows:

\begin{align}  
    \tilde{\boldsymbol{\ell}}^{i:} = \mathcal{J(\boldsymbol{\ell^{i:}}; \varphi)} \in & \mathbb{R}^{K}, \notag \quad   \Psi_i(d_t, \phi) = (\mu^{i:}, \sigma^{i:^2}) \in \mathbb{R}^{2\times K},    \\
      \mathcal{R}^{i} &= - \log_2(\mathcal{N}(\tilde{\boldsymbol{\ell}}^{i:}|\mu^{i:}, \sigma^{i:^2} )) \label{grem_jumper},
\end{align}

\noindent where $ \tilde{\boldsymbol{\ell}}^{i:}$ is the projection performed by $\mathcal{J}^i$ to a representation of dimension $K\leq N$; this dimensionality reduction determines the number of parameters required for the EE module. 
We point out that each EE employs a distinct module; however, this does not pose a significant issue, as the cumulative parameter count for these modules remains negligible compared to the overall parameter size of the combined image encoder.

To further improve final discrimination capabilities of these  modules, we can increase the size of $K$ until $ \tilde{\boldsymbol{\ell}}^{i:}$  matches the dimension of $d_t$, as performed by the last projection layer of CLIP. 
This way we can add  to eq. \ref{loss} the cross-modal similarity between the two latent space, obtaining the final loss function:
\begin{align} \label{final_loss}
    \mathcal{L} = \mathcal{L}_{r} &+  \mathcal{L}_{s},  \\
    \mathcal{L}_{r} =     -\mathbb{E} [ \log_2(\mathcal{N}(\tilde{\boldsymbol{\ell}}^{i:}|\mu^{i:}, \sigma^{i:^2}) ], &\quad \mathcal{L}_{s} = 1 - CS(d_{t},\tilde{\boldsymbol{\ell}}^{i:}), \notag
\end{align}

\noindent where $CS$ stands for cosine-similarity function; the loss pipeline is depicted in fig. \ref{fig3a}.
To summarize, the \emph{class-rate} $\mathcal{L}_{r}$ serves as a regularization term that enhances the ability to accurately discriminate between different semantic contents.
Architecture of both $\mathcal{J}^i$ and $\Psi^i$ are  shown on fig. \ref{fig3a}.

During inference, this architecture can be utilized in a manner similar to that described in Sect.  \ref{sb}, with the mean and variance extracted with $\Psi^i$.

\section{Experiments}
\vspace{-0.25cm}
This section evaluates the T-GEM modules and the \emph{class-rate} impact on learning, considering classification to be our test-case scenario.
Section \ref{ts} details setup and training, while Section \ref{pr} compares T-GEMs and sample-based methods against Full CLIP, considering different setups and also evaluating the gains in terms of number of parameters. 
In Section \ref{ila}  we further analyzed what happens with respect to different EE layers and the justifications why computing the rate can lead to a better discrimination across classes.

\subsection{Training setup}\label{ts}
For the sampling-based approach, we select 100 samples per target class and calculate the mean and standard deviation. 

For the learning-rate approach, we consider  the vision transformer based CLIP ViT-B-32 and ViT-L-14 \cite{Radford2021LearningTV}; their image encoder is organized into 12 transformer residual blocks (RBs) for a total of  around 86 and 300 millions parameters.
In our experiments both $\Psi^i$ and $\mathcal{J}^i$ are simply MLP, and  we consider the right $K$ in order to be able to compute cosine similarity.
We train both $\Psi^i$ and $\mathcal{J}^i$ optimizing eq. \ref{final_loss}, considering CIFAR10 \cite{krizhevsky2009learning} and one text description for each class.
Since this work is meant to be a first exploration toward rate-based early-exit methods, we introduce EE models after each of the RB of the image encoder, and the results will be represented with RB$i$, where $i$ is the intermediate early exit layer.
We used Adam optimizer \cite{Kingma2015adam}  for 120 epochs, decreasing the initial learning rate of 0.0001 on a plateau.

\subsection{Classification results} \label{pr}
Tab. \ref{tab1}  shows results considering 1000 test samples uniformly distributed over different classes, considering VIT-B-32 and VIT-L-14; results are separated by the symbol $/$ in the table.
We divide results in two groups: sampling-based and learning-based approaches.
Concerning the former, we evaluate our class-rate against using the average cosine similarity between the test sample and $\boldsymbol{\mathcal{S}}$, a fair baseline and the only alternative for EE without the text encoder. 
Our results show consistent performance gains across almost all RBs and for both models, indicating that intermediate probability distribution of the activation maps carry more informative signals for EE than simple cosine similarity.

Regarding the learning-based methods, best results are obtained when still considering \emph{Cosine similarity} at inference.
However, two critical observations emerge; first, incorporating $\mathcal{L}_{r}$ as regularizer in the loss functions enhances the model's ability to project intermediate latent representations into a more discriminative space, resulting in improved performance.
This improvement is particularly noticeable from RB3 to RB5, where the gain reaches up to 6\%.
Second, the class-rate metric demonstrates comparable results, indicating its effectiveness as a reliable proxy for cosine similarity.

\begin{figure}[h]
    \centering
    \includegraphics[width=0.8\columnwidth]{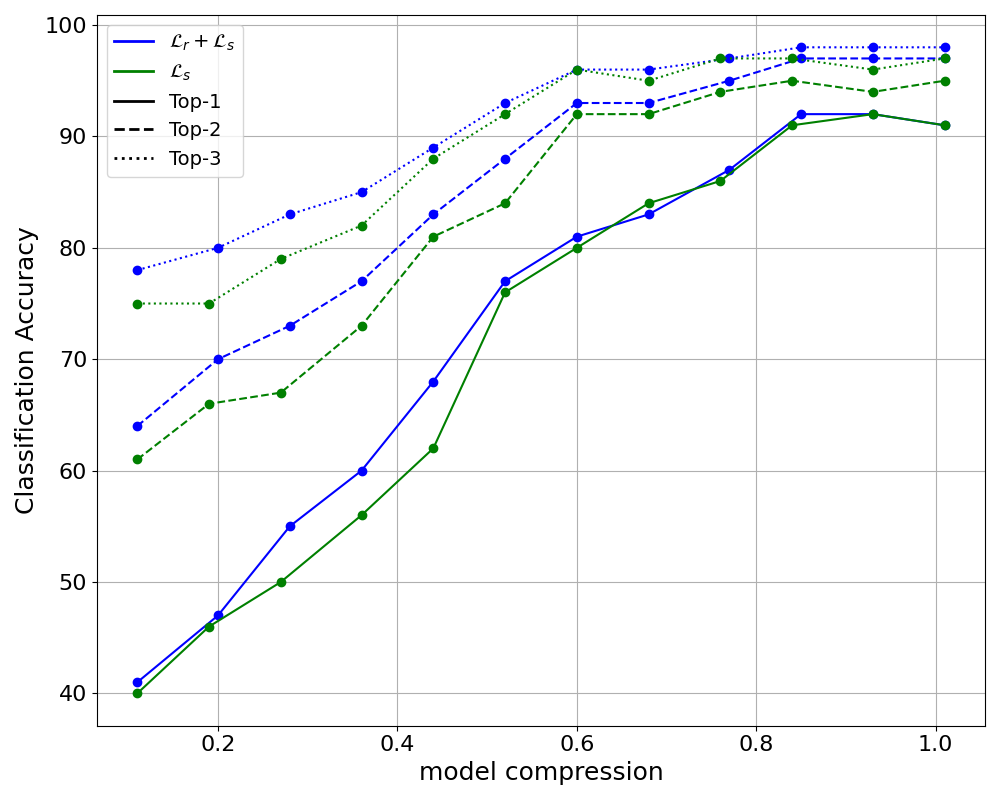}
    \vspace{-0.5cm}
    \caption{Model compression (the lower the better) vs top-$\{1,2,3\}$ accuracy (the higher the better).}
    \label{fig3}
\end{figure}

Tab. \ref{tab1} shows that both $\Psi^i$ and $\mathcal{J}^i$ have only a few hundred thousand parameters, which is about 4\% of a RB when considering ViT-B-32.
This means that, for example, if we want to add an EE after the fifth layer, we can obtain an equivalent encoder with around 52 million fewer parameters.
Furthermore,  fig. \ref{fig3} shows model compression vs. top-$\{1,2,3\}$ accuracy, where model compression represents the ratio between the number of parameters with the EE at different RBs and the total parameters of the image encoder. 
We compare EE modules trained with and without the class-rate regularizer, observing  that the latter improves classification accuracy for most model compression level, especially for top-$\{2,3\}$.
Observing top-$\{2,3\}$, we see that even when our method fails to select the correct class, it identifies plausible alternatives; this can have interesting application in field like clustering  or conformal prediction \cite{angelopoulos2021gentle}.

\begin{table}[ht]
\centering
\caption{Results when varying $K$ in the learning-based approach and the size of $\boldsymbol{\mathcal{S}}$ in the sampling-based approach }\label{tab2}
\resizebox{\columnwidth}{!}{%
\begin{tabular}{l cc ccc}
\toprule
\textbf{Method} &   \textbf{Setting} &   \textbf{AP}  &
\multicolumn{3}{c}{\textbf{CIFAR10}}  \\
\cmidrule(lr){4-6} 
& &  & RB2 & RB4 & RB6     \\
\midrule
Sampling-based & \#($\boldsymbol{\mathcal{S}}$) = 1  & 0  &  18 & 18 & 20    \\
Sampling-based & \#($\boldsymbol{\mathcal{S}}$) = 10  & 0  &  29 & 35  & 48    \\
Sampling-based & \#($\boldsymbol{\mathcal{S}}$) = 100  & 0 &   32 & 41  & 57  \\
Sampling-based & \#($\boldsymbol{\mathcal{S}}$) = 250  & 0 &   34 & 42 & 58  \\
Sampling-based & \#($\boldsymbol{\mathcal{S}}$) = 1000  & 0 &  33 & 43  & 59  \\
\bottomrule
T-GEMs  + Jumper  & K = 16  &  46.6k & 17 &13 & 20  \\
T-GEMs  + Jumper  & K = 32  &  91.7k & 42 & 51 & 75 \\
T-GEMs  + Jumper  & K = 128   &  362k & 47 & 60 & 77   \\
T-GEMs + Jumper  & K = 512  & 1.45M &  47 & 59 & 77  \\
T-GEMs + Jumper  & K = 1024  & 2.9M & 47 & 60 & 74  \\
\bottomrule
\end{tabular}%
}

\end{table}

\subsection{Varying model complexity and qualitative analysis} \label{ila}

Table \ref{tab2} illustrates the variations in accuracy as the number of samples per class is adjusted for extracting distribution parameters in both sampling-based and learning-based scenarios.
Specifically, it examines the impact of altering the sample count for the former (\#($\boldsymbol{\mathcal{S}}$)) and modifying the parameter $K$ for the latter, considering VIT-32-B.
Concerning the sampling-based, it is necessary to have a reasonable amount of samples for each class to obtain reliable results; when $\boldsymbol{\mathcal{S}} = 1$, accuracy does not improve when going toward the RBs, showing that we need more images.
On the other hand we also show that it is not necessary to have a huge amount of samples, since results for $\boldsymbol{\mathcal{S}} \geq 100$ are similar. 
Concerning the learning-based approach, the number of additional parameters (AP in tab. \ref{tab2}) increases with K, going beyond one million for $K \geq 512$. However, performance stabilizes on $K = 128$, showing that it is not necessary to have large models.

EE methods optimize exit layer selection per sample, leveraging the intuition that full processing is not always necessary.
Our approach does not tackle this task but demonstrates: 1) Intermediate activations distribution, even in early encoder layers, can be discriminative, and 2) Incorporating rate as a regularizer during training enhances discrimination.
To prove this we conducted a preliminary experiment, aided by an Oracle (not a real EE method) that indicates classification correctness.
Starting from the first RB, the EE progresses to subsequent blocks until the correct class is identified or the encoder ends. This approach leads to 94\% accuracy, where 53.6 \%  of the time the optimal EE is before the 6-th RB.

\section{Conclusion and future direction}
\vspace{-0.3cm}
We introduced a reliable approach to leverage intermediate layers in MM models, such as CLIP, utilizing the computation of activation rates across subsequent layers.
Our method based on T-GEMs and the Jumper achieves robust performance, reducing the size of the image encoder without compromising results in terms of image-text similarity, achieving a firt promising static EE method. 
Furthermore, the introduction of $ \mathcal{L}_{r}$ in eq. \ref{final_loss} improves accuracy of the EE modules, showing its potential as a regularizer to extract more discriminative representations,  yielding superior outcomes.
The proposed approach can be further improved, by for instance converting it to be dynamic; a confidence score can be computed for each sample, enabling the selection of the optimal intermediate layer from which to exit.
Furthermore, the proposed rate-based EE method can also be used to fine-tune or train a lightweight MM encoder.

\bibliographystyle{IEEEbib}
\bibliography{strings,refs}

\end{document}